\documentclass[11pt]{article}

\usepackage{times}
\usepackage{inconsolata}
\usepackage{microtype}
\usepackage{textcomp}

\usepackage{amsmath}
\usepackage{amssymb}
\usepackage{amsthm}
\usepackage{mathrsfs}

\usepackage{graphicx}
\usepackage{svg}
\usepackage{booktabs}

\usepackage{lipsum}
\usepackage{cite}
\usepackage{url}
\usepackage{pifont}
\usepackage{tikz}
\usepackage{paralist}
\usepackage{multirow}
\usepackage{xifthen}
\usepackage{enumerate}
\usepackage{titlesec}
\usepackage{color}
\usepackage{bm}
\usepackage{caption}
\usepackage{subcaption}
\usepackage{soul}
\usepackage{xspace}
\usepackage{authblk}

\usepackage{algorithm}
\usepackage{algpseudocode}
\usepackage{adjustbox}
\usepackage{colortbl}
\usepackage{dblfloatfix}
\usepackage{multicol}
\usepackage{tcolorbox}
\tcbuselibrary{breakable,skins,theorems}
\usepackage{xparse}
\usepackage{hyperref}
\usepackage{cleveref}
\usepackage{capt-of}
\usepackage{etoolbox}
\usepackage{listings}
\usepackage{shadowtext}
\usepackage{array}

\pdfoutput=1

\usepackage{acl}

\DeclareRobustCommand{\method}{\textsc{DeepPresenter}\xspace}
\DeclareRobustCommand{\model}{DeepPresenter-9B\xspace}

\DeclareRobustCommand{\researcher}{Researcher\xspace}
\DeclareRobustCommand{\presenter}{Presenter\xspace}
\DeclareRobustCommand{\inspect}{\texttt{inspect}\xspace}
\DeclareRobustCommand{\think}{\texttt{think}\xspace}

\makeatletter

\title{\textsc{DeepPresenter}: Environment-Grounded Reflection for \\  Agentic Presentation Generation}

\author{
  \textbf{Hao Zheng}${}^{1,2,*}$,
  \textbf{Guozhao Mo}${}^{1,2,*}$,
  \textbf{Xinru Yan}${}^{2}$,
  \textbf{Qianhao Yuan}${}^{1,2}$,
  \textbf{Wenkai Zhang}${}^{1}$,\\
  \textbf{Xuanang Chen}${}^{1}$,
  \textbf{Yaojie Lu}${}^{1,\dagger}$,
  \textbf{Hongyu Lin}${}^{1}$,
  \textbf{Xianpei Han}${}^{1}$,
  \textbf{Le Sun}${}^{1}$ \\
  ${}^{1}$Chinese Information Processing Laboratory, Institute of Software,\\
  Chinese Academy of Sciences, Beijing, China\\
  ${}^{2}$University of Chinese Academy of Sciences, Beijing, China\\
  \texttt{\{zhenghao2022,moguozhao2024,chenxuanang,luyaojie,hongyu\}@iscas.ac.cn}
  
}

\begin{document}
\maketitle
  \renewcommand{\thefootnote}{\fnsymbol{footnote}}
\footnotetext[1]{These authors contributed equally to this work.}
\footnotetext[2]{Corresponding authors.}
\renewcommand{\thefootnote}{\arabic{footnote}}
\begin{abstract}
    Presentation generation requires deep content research, coherent visual design, and iterative refinement based on observation. 
    However, existing presentation agents often rely on predefined workflows and fixed templates. 
    To address this, we present \method, an agentic framework that adapts to diverse user intents, enables effective feedback-driven refinement, and generalizes beyond a scripted pipeline.
    Specifically, \method autonomously plans, renders, and revises intermediate slide artifacts to support long-horizon refinement with environmental observations.
    Furthermore, rather than relying on self-reflection over internal signals (e.g., reasoning traces), our environment-grounded reflection conditions the generation process on perceptual artifact states (e.g., rendered slides), enabling the system to identify and correct presentation-specific issues during execution.
    Results on the evaluation set covering diverse presentation-generation scenarios show that \method achieves state-of-the-art performance, and the fine-tuned \model remains highly competitive at substantially lower cost. Our project is available at: \url{https://github.com/icip-cas/PPTAgent}
\end{abstract}
\begin{figure}[ht!]
  \centering
  \includegraphics[width=1.0\linewidth]{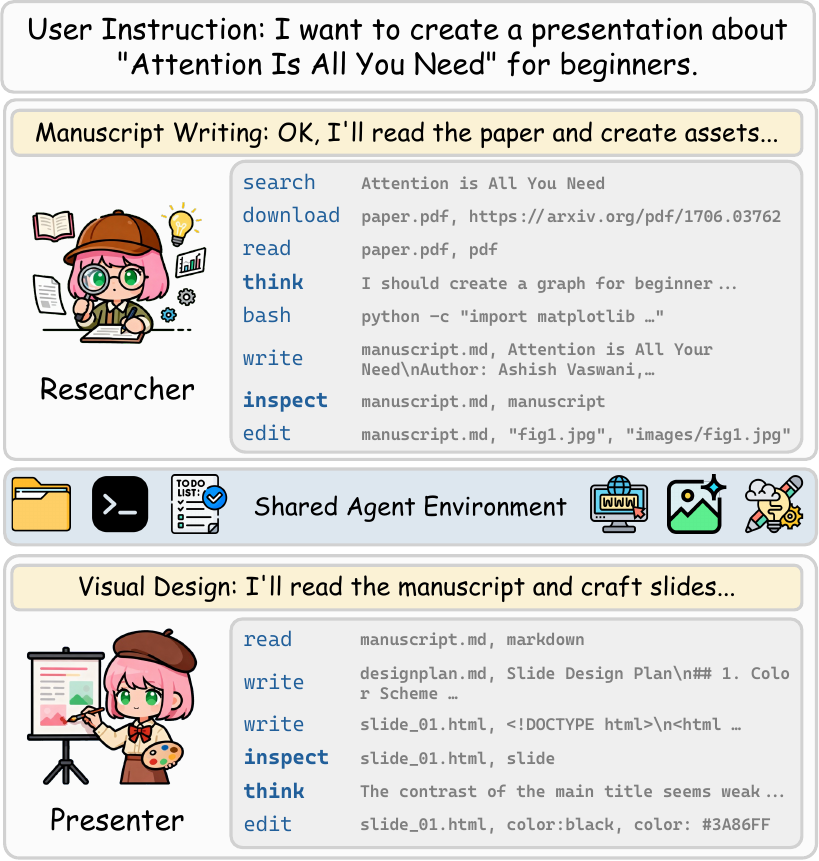}
  \caption{Illustration of \method. Given a user instruction, the \researcher gathers information and compiles a structured manuscript, while the \presenter transforms it into visual slides. Both agents interact and collaborate with a shared environment, leveraging grounded observations for reflective refinement.}
  \label{fig:1}
  \vspace{-8pt}
\end{figure}

\section{Introduction}
Presentations are a primary medium for information delivery across education, business, and research.
A high-quality presentation combines well-researched content with coherent visual design, enabling audiences to grasp complex ideas efficiently.
However, creating such presentations remains time-consuming and skill-demanding, motivating recent work that leverages Multimodal Large Language Models (MLLMs) to automate this task \citep{liang2025slidegen,zheng2025pptagent,yang2025auto}.

However, existing presentation agents \citep{sefid2021slidegen,xu2025pregenie,yang2025auto} fall short of meeting these demands.
First, they rely on predefined workflows \citep{zheng2025pptagent} and content-agnostic templates \citep{cachola-etal-2024-knowledge}, limiting adaptability to varying user intents. This yields text-heavy slides with insufficient research depth and visual designs that fail to resonate with the narrative.
Second, introspective reflection over internal signals (e.g., code or reasoning traces) cannot detect post-render defects \citep{tang2025realcritic,kim2025reflact}, resulting in overlapping elements, truncated text, and broken layouts.

To address these limitations, we propose \method, an agentic framework for presentation generation (Figure~\ref{fig:1}).
Unlike prior methods that decouple content and design via rigid templates, \method coordinates two specialized agents through a shared observation space.
The \researcher autonomously explores and compiles a structured manuscript aligned with the user intent, while the \presenter converts it into visually coherent slides via content-driven design rather than template filling.
Crucially, instead of introspective self-reflection over internal signals, \method grounds reflection in perceptual artifact states obtained from environmental observation (Figure~\ref{fig:reflection}): agents use \inspect to view rendered manuscripts and slides, and \think to plan targeted revisions to correct post-render defects.

While our framework achieves strong performance with proprietary models, their high cost motivates a more efficient alternative. We therefore develop \model via supervised fine-tuning on curated trajectories (Figure~\ref{fig:method}).
We first construct diverse presentation tasks from PersonaHub \citep{ge2024personahub}, arXiv, and FinePDFs \citep{kydlicek2025finepdfs}, augmented with verifiable constraints.
During trajectory synthesis, we mitigate self-verification bias \citep{stechly2024self} with extrinsic verification: an independent critic evaluates artifacts in isolation and provides reasoning traces that steer targeted refinements, improving the quality of synthesized trajectories.

We evaluate our method on a held-out set of 128 diverse presentation tasks across three dimensions: constraint satisfaction, content quality, and visual style.
With proprietary backbones, \method achieves an average score of 4.44, surpassing open-source baselines and the commercial system Gamma (4.36).
Our specialized agentic design yields richer content and coherent design, while environment-grounded reflection reduces post-render defects by revising against observed perceptual artifact states.
\model scores 4.19, outperforming all open-source baselines and approaching GPT-5 (4.22) at lower cost.

In summary, our contributions are threefold:
\begin{itemize}

  \item We propose \method, an agentic presentation framework that coordinates \researcher and \presenter agents via a shared observation space, enabling autonomous information research and topic-aware design.

  \item We introduce environment-grounded reflection that grounds self-correction in perceptual artifact states obtained from post-render observations, reducing defects that are not detectable from internal signals alone.

  \item Results on the evaluation set covering diverse presentation-generation scenarios show that \method achieves state-of-the-art performance, and the distilled \model remains highly competitive at substantially lower cost.

\end{itemize}

\section{\method}

\begin{figure}[t!]
  \centering
  \includegraphics[width=1.0\linewidth]{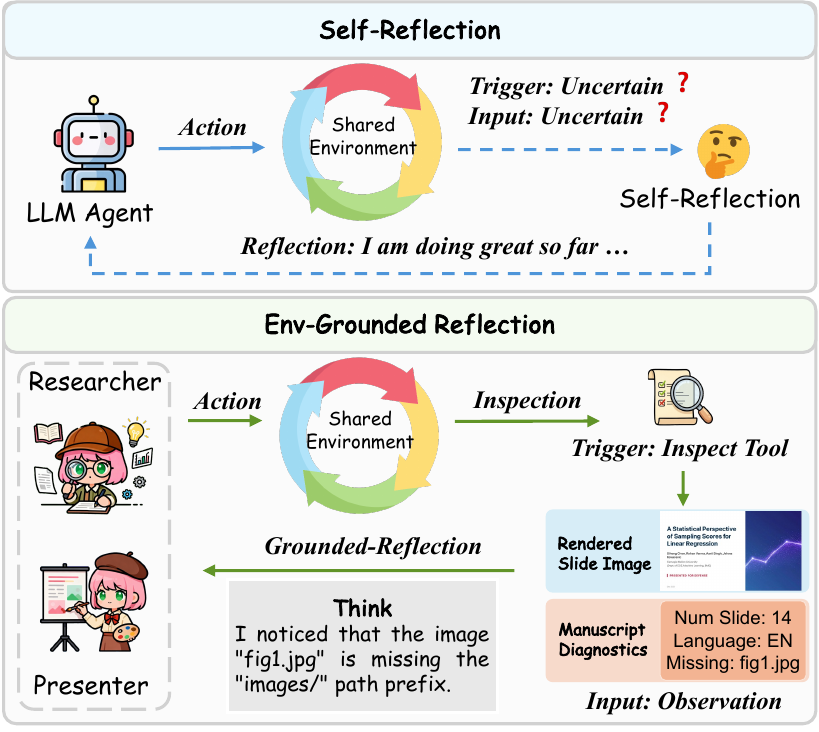}
  \caption{Comparison between self-reflection and environment-grounded reflection. Self-reflection relies on uncertain triggers and inputs without external signals. \method grounds reflection in environmental observations through the \inspect tool.}
  \label{fig:reflection}
  \vspace{-5pt}
  
\end{figure}

In this section, we present \method, a dual-agent framework for presentation generation.
We first formulate the task as an interactive agentic process, then
describe the \researcher-\presenter collaboration and the environment-grounded
reflection mechanism, as illustrated in Figure~\ref{fig:reflection}.

\subsection{Task Formulation}
\label{sec:formulation}

We formulate presentation generation as an interactive agentic task.
Given an instruction $\mathcal{I}$ and an agent environment $\mathcal{E}$
equipped with a tool library $\mathcal{T}$ and a file system $\mathcal{F}$, the system aims to
generate a high-quality presentation $\mathcal{P}$. 
The generation process can be modeled as a
multi-step trajectory $\tau = \{(r_1, a_1, o_1), \dots, (r_T, a_T, o_T)\}$,
where at each step $t$, the agent generates a reasoning trace $r_t$,
selects an action $a_t \in \mathcal{T}$, and receives observation $o_t$
from $\mathcal{E}$. We decompose the trajectory into two sequential phases:
$\tau = \tau^R \circ \tau^P$, where $\tau^R$ and $\tau^P$ denote the
\researcher and \presenter trajectories, respectively. The two agents
communicate through $\mathcal{F}$,
where the \researcher persists a structured manuscript $\mathcal{M}$ and
associated assets for the \presenter to consume. Appendix~\ref{appendix:agent} lists the tools.

\subsection{Dual-Agent Collaboration}
\label{sec:collaboration}

Presentation generation requires both information research and visual design, which demand different planning and tool use. We split these roles between two specialized agents while sharing the same backbone model.

\paragraph{Researcher Agent}

Given $\mathcal{I}$, the \researcher autonomously plans its exploration instead of following a predefined workflow. It executes multiple steps during $\tau^R$, invoking tools from $\mathcal{T}$ to retrieve and synthesize supporting materials and to create auxiliary assets as needed.
The exploration depth and strategy adapt to user intent: a technical presentation may require surveying related work, while a general-audience talk may prioritize accessible examples and vivid illustrations.
Finally, the \researcher compiles slide text and associated assets into a structured markdown manuscript $\mathcal{M}$ organized by narrative flow, and persists it to $\mathcal{F}$.

\begin{table}[t]
  \centering
  \resizebox{\linewidth}{!}{
    \begin{tabular}{lccc}
      \toprule
      \textbf{Dimension} & \textbf{Category} & \textbf{Count} & \textbf{Ratio (\%)} \\
      \midrule
      \multirow{2}{*}{Language} & English & 603 & 52.34 \\
      & Chinese & 549 & 47.66 \\
      \midrule
      \multirow{3}{*}{Source} & PersonaHub & 586 & 50.87 \\
      & FinePDFs & 362 & 31.42 \\
      & arXiv & 204 & 17.71 \\
      \midrule
      \multirow{4}{*}{Aspect Ratio} & 16:9 Widescreen & 327 & 28.39 \\
      & 4:3 Standard & 304 & 26.39 \\
      & A1 Poster & 30 & 2.60 \\
      & Free & 491 & 42.62 \\
      \midrule
      \multirow{3}{*}{Slide Count} & 11-20 & 249 & 21.61 \\
      & 1-10 & 320 & 27.78 \\
      & Free & 583 & 50.61 \\
      \midrule
      \multicolumn{2}{l}{\textbf{Total}} & \textbf{1,152} & \textbf{100.00} \\
      \bottomrule
  \end{tabular}}
  \caption{Statistics of the constructed presentation tasks by language, source, aspect ratio, and slide count. ``Free'' indicates no constraint is specified.}
  \label{table:data_stat}
  \vspace{-5pt}
\end{table}

\paragraph{Presenter Agent}
Rather than populating predefined templates, the \presenter generates slides from scratch during $\tau^P$. Given $\mathcal{M}$ from $\mathcal{F}$, the agent first develops a global design plan, establishing color themes and typography that resonate with the topic.
It then generates each slide as a standalone HTML file, translating manuscript content into visual elements following the design plan. This content-driven approach enables stylistic
choices aligned with the presentation topic, such as earthy palettes for sustainability or minimalist layouts for academic tutorials.

\subsection{Environment-Grounded Reflection}
\label{sec:reflection}

We ground agent reflection in environmental observations rather than introspective reasoning over internal signals \citep{he-etal-2025-large}. The key issue with self-reflection is state mismatch: agents operate on intermediate representations (e.g., HTML or markdown), while users perceive only rendered artifacts. As a result, many defects manifest only in perceptual states (e.g., broken images, overflow, or low contrast), leaving introspective reflection operating in a mismatched observation space.

To make perceptual artifact states observable to the agent, we introduce the \inspect tool as an explicit observation interface. For the \presenter, \inspect renders an HTML file into image pixels, exposing post-render defects such as overflow, overlap, and low contrast; for the \researcher, \inspect returns structured diagnostics of the manuscript and file state, including slide count, asset availability, and detected language. 
Agents then use \think to reflect on observed defects and plan targeted edits. This forms an observe--reflect--revise loop where agent observations align with user perception.

\begin{figure*}[t]
  \centering
  \includegraphics[width=\linewidth]{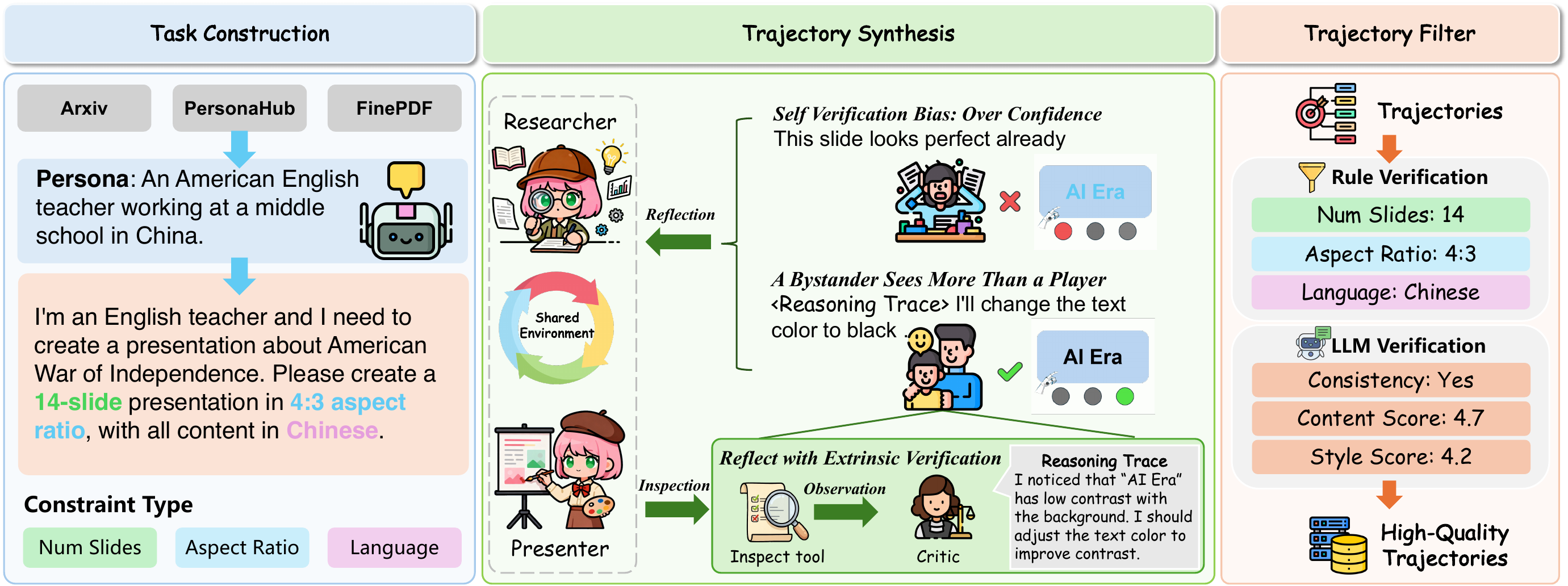}
  \caption{Our data synthesis pipeline. The process ensures high-quality
    trajectories for supervised fine-tuning through three integrated mechanisms:
    (1) Query Construction augments tasks with verifiable constraints;
    (2) Extrinsic Verification injects reasoning traces when defects are identified to guide
    agent self-correction during sampling; and (3) Trajectory Filtering
  validates constraint compliance and assesses consistency and output quality.}
  \label{fig:method}
\end{figure*}
  \vspace{-5pt}

\section{Frontier Presentation Agent Model}

This section presents our training pipeline as shown in Figure~\ref{fig:method}: task dataset construction, trajectory synthesis with extrinsic verification to elicit high-quality reflective behaviors, and multi-stage filtering for quality.

\label{sec:training}

\subsection{Query Construction}
We construct a task collection for training our compact model and evaluating our framework. To cover diverse presentation scenarios in both intent-driven and document-conditioned settings, we draw task seeds from PersonaHub \citep{ge2024personahub}, arXiv, and FinePDFs-Edu \citep{kydlicek2025finepdfs}. Each task is augmented with verifiable constraints (e.g., slide count, language, aspect ratio) to capture fine-grained user-specified requirements.
For PersonaHub, we prompt GLM-4.6 to synthesize presentation tasks conditioned on persona descriptions; for arXiv and FinePDFs-Edu, we construct tasks that require generating presentations based on provided documents.
Each task is further augmented with verifiable constraints, including slide count, language, and aspect ratio.
In total, this task collection contains 1,152 tasks, with 1,024 for trajectory sampling and 128 held out for evaluation. Detailed statistics are shown in Table~\ref{table:data_stat}.

\subsection{Verification-Guided Trajectory Synthesis}
When sampling agentic trajectories, self-reflection is susceptible to self-verification bias~\citep{jiang2025self}: the agent judges its own intermediate outputs from within the same trajectory state that produced them. This coupling entangles verification with self-justification, resulting in flawed outputs being accepted. To break this coupling, we introduce extrinsic verification, where verification signals are produced in an isolated context.

As illustrated in Figure~\ref{fig:method}, after the agent invokes \inspect and obtains an observation $o_t$, an independent critic performs verification conditioned on $o_t$ and the corresponding intermediate artifacts. The critic outputs a reasoning trace that identifies defects (e.g., low contrast) and specifies actionable adjustments (e.g., adjust text color). We append this trace to the agent context as a \think call, guiding targeted revisions before continuing the rollout.

\begin{table*}[ht!]
  \centering
  \begin{tabular}{llccccc}
    \toprule
    \textbf{Framework} & \textbf{Model} & \textbf{Constraint} & \textbf{Content} & \textbf{Style} & \textbf{Avg.} & \textbf{Diversity} \\
    \midrule

    \rowcolor{gray!20} \multicolumn{7}{c}{\textit{Close-sourced Baseline}} \\
    Gamma & -- & \underline{4.93} & \underline{4.08} & 4.08 & 4.36 & 0.52 \\
    \midrule

    \rowcolor{gray!20} \multicolumn{7}{c}
    {\textit{Open-sourced Baseline}} \\
    \multirow{4}{*}{PPTAgent} & GPT-5 & 3.96 & 3.00 & 4.07 & 3.68 & 0.35 \\
    & Gemini-3-Pro & 4.22 & 3.09 & \underline{4.30} & 3.87 & 0.19 \\
    & Claude-Sonnet-4.5 & 3.72 & 2.93 & 4.15 & 3.60 & 0.17 \\
    & GLM-4.6 & 4.02 & 3.17 & 4.24 & 3.81 & 0.30 \\
    \cmidrule(lr){1-7}
    \multirow{4}{*}{KCTV} & GPT-5 & \textbf{4.95} & 2.84 & 3.63 & 3.81 & 0.21 \\
    & Gemini-3-Pro & 4.58 & 3.01 & 3.90 & 3.83 & 0.27 \\
    & Claude-Sonnet-4.5 & 4.88 & 2.90 & 3.99 & 3.92 & 0.20 \\
    & GLM-4.6 & 4.66 & 2.83 & 3.94 & 3.81 & 0.25 \\

    \midrule
    \rowcolor{gray!20} \multicolumn{7}{c}{\textit{Ours}} \\
    \multirow{6}{*}{\method} & GPT-5 & 4.80 & 3.79 & 4.07 & 4.22 & 0.56 \\
    & Gemini-3-Pro & 4.70 & \textbf{4.25} & \textbf{4.37} & \textbf{4.44} & \textbf{0.79} \\
    & Claude-Sonnet-4.5 & 4.90 & 4.05 & 4.27 & \underline{4.41} & 0.49 \\
    & GLM-4.6V & 4.69 & 3.25 & 3.75 & 3.90 & \underline{0.58} \\
    & GLM-4.6V-Flash & 4.67 & 3.11 & 3.69 & 3.82 & 0.47 \\
    & \model & 4.77 & 3.52 & 4.29 & 4.19 & 0.53\\

    \bottomrule
  \end{tabular}
  \caption{Performance comparison of different frameworks and models. The best/second-best scores are \textbf{bolded}/\underline{underlined}. Quality metrics (Constraint, Content, Style, Avg.) are scaled to 0–5, while Diversity is scaled to 0–1.}
  \label{tab:comparison}
\end{table*}

\subsection{Trajectory Filtering}
\label{sec:filter}
We adopt a three-stage filtering pipeline to ensure trajectory quality.
First, we verify \textit{constraint compliance} through a rule-based system.
Second, we evaluate \textit{consistency} using GLM-4.6, removing trajectories that fail to follow the extrinsic-verification trace with aligned revisions (i.e., reflection--action inconsistency).
Third, we assess \textit{output quality} using GLM-4.6V, filtering out trajectories with critical defects such as element overlap or broken images.

\section{Experiment}
\label{sec:exp}

In this section, we evaluate our method on presentation generation and analyze
our key components.

\subsection{Setup}

\paragraph{Implementation Details}
We sample trajectories by running \method with Gemini-3-Pro as the backbone and critic model on 1,024 training tasks, with a maximum context window of 50K tokens.
802 trajectories pass our filtering pipeline and are used for supervised fine-tuning. We fine-tune GLM-4.6V-Flash on these trajectories using MS-SWIFT \citep{zhao2024swiftascalablelightweightinfrastructure}, with a batch size of 32 and learning rate of 1e-5 for 5 epochs. Training takes approximately 80 GPU hours on 8 A800 GPUs.

\paragraph{Models and Baselines}
We compare against one commercial system, Gamma\footnote{https://gamma.app/}, and two academic frameworks: PPTAgent \citep{zheng2025pptagent} and KCTV \citep{cachola-etal-2024-knowledge}. For backbone models, we evaluate with proprietary GPT-5 \citep{openaiIntroducingGPT5}, Gemini-3-Pro \citep{comanici2025gemini25pushingfrontier}, and Claude-Sonnet-4.5 \citep{anthropicIntroducingClaudeSonnet4.5}, as well as open-source GLM-4.6 \citep{zeng2025glm}. For \method, we additionally evaluate with GLM-4.6V and GLM-4.6V-Flash \citep{vteam2025glm45vglm41vthinkingversatilemultimodal}, as our framework leverages visual feedback through the \inspect tool.

\paragraph{Evaluation Protocol}
We hold out 128 tasks from the constructed task collection and evaluate generated presentations using the following metrics:

$\bullet$ \textbf{Constraint} scores each presentation by the fraction of user-specified constraints it satisfies, covering slide count, language, and aspect ratio, verified through rule-based checking.

$\bullet$ \textbf{Content \& Style} evaluate the quality of slide content and visual design. We adopt the MLLM-based evaluation framework from \citet{zheng2025pptagent} with GPT-5 as the judge, which has been validated to correlate well with human judgments.

$\bullet$ \textbf{Diversity} quantifies visual style variance across generated presentations using the Vendi Score \citep{friedman2022vendi}, which computes diversity based on the eigenvalue entropy of feature similarity matrices extracted by DINOv2 \citep{oquab2023dinov2}.

We report Avg. as the mean of Constraint, Content, and Style (scaled 0--5), while Diversity (scaled 0--1) measures cross-presentation variation.

\subsection{Main Results}

Table~\ref{tab:comparison} presents the main experimental results.

\paragraph{\method achieves state-of-the-art performance}
Across all backbone models, \method consistently outperforms open-source baselines. With Gemini-3-Pro as the backbone, \method attains an average score of 4.44, surpassing the best open-source baseline (KCTV + Claude-Sonnet-4.5, 3.92) by 13.3\% and the commercial product Gamma (4.36).
The improvements stem from two aspects:
(1) 
\textit{Content quality improves most because \researcher performs intent-adaptive information seeking and synthesis, rather than relying on fixed workflows or user-provided inputs.}
Baseline frameworks depend on user-provided materials and lack deep retrieval capability, while our agent searches, retrieves, and synthesizes information from diverse sources.
(2) \textit{Style scores improve through content-aware design and environment-grounded reflection.} Our framework enables \presenter to align design decisions with the narrative, while environment-grounded reflection mitigates free-form generation failures by revising against post-render defects.

\paragraph{Free-form generation enables greater visual diversity, with \method achieving a diversity score of 0.79.}
Under our diversity metric, \method more than doubles template-based baselines by generating slides in a free-form manner. Baseline frameworks achieve diversity scores of only 0.17 to 0.35, as fixed templates constrain visual variation. 
PPTAgent, in particular, shows lower constraint scores because its style decisions are predetermined by the workflow, limiting task-specific adaptation.
Even Gamma, despite its commercial polish, achieves only 0.52. In contrast, our framework maintains high constraint compliance while enabling greater visual diversity (0.79).

\paragraph{\model surpasses all open-source baselines with high efficiency.}
With only 802 trajectories, our compact model achieves an average score of 4.19, outperforming open-source baselines and matching GPT-5 (4.22) at substantially lower cost. These results support the effectiveness of our verification-guided trajectory synthesis and suggest that compact models can acquire agentic behaviors from limited but high-quality samples.

\begin{table}[t!]
  \centering
  \resizebox{\columnwidth}{!}{
    \begin{tabular}{lcccc}
      \toprule
      \textbf{Configuration} & \textbf{Cons.} & \textbf{Content} & \textbf{Style} & \textbf{Avg.} \\
      \midrule
      \rowcolor{gray!20} \multicolumn{5}{c}{\textit{Gemini-3-Pro}} \\
      \method            & 4.70 & 4.25 & 4.37 & 4.44 \\
      w/o Grounded Reflection    & 4.52 & 4.15 & 4.31 & 4.32 \\
      w/o Dual-Agent       & 3.94 & 3.96 & 4.22 & 4.04 \\
      \midrule
      \rowcolor{gray!20} \multicolumn{5}{c}{\textit{DeepPresenter-9B}} \\
      \model & 4.77 & 3.52 & 4.29 & 4.19\\
      w/o Grounded Reflection  &  4.21 & 3.23 & 4.01 & 3.82 \\
      w/o Dual-Agent    &  3.65 & 2.93 & 3.11 & 3.23\\
      w/o Trajectory Filtering   &  4.67 & 3.30 & 4.12 & 4.03 \\
      \bottomrule
  \end{tabular}
  }
  \caption{Ablation study on framework components and training strategy. Cons. denotes constraint satisfaction.}
  \label{tab:ablation}
\end{table}

\begin{table}[t!]
  \centering
  \resizebox{\columnwidth}{!}{
    \begin{tabular}{lcccccc}
      \toprule
      \textbf{Configuration} & \textbf{Cons.} & \textbf{Content} & \textbf{Style} & \textbf{Avg.} & \textbf{$\Delta$} \\
      \midrule
      GLM-4.6V-Flash & 4.67 & 3.11 & 3.69 & 3.82 & -- \\
      \midrule
      + Fine-tuning & 4.71 & 3.19 & 3.92 & 3.94 & +0.12 \\
      + Extrinsic Verification & 4.74 & 3.28 & 4.03 & 4.02 & +0.20 \\
      \bottomrule
    \end{tabular}
  }
  \caption{Effect of extrinsic verification on model performance. Both fine-tuned variants use 300 trajectories. $\Delta$ denotes improvement over the base model.}
  \label{tab:process_oversight}
\end{table}

\subsection{Ablation Study}
We ablate key components of \method on Gemini-3-Pro and \model, as shown in Table~\ref{tab:ablation}.
(1) \textit{Environment-grounded reflection is critical because it extends observation space to post-render perceptual artifact states.} Disabling \inspect confines reflection to pre-render artifacts and degrades performance from 4.44 to 4.32 on Gemini-3-Pro and from 4.19 to 3.82 on \model.
(2) \textit{Dual-agent collaboration contributes significantly by decomposing long-horizon execution into specialized sub-tasks.} Without it, performance drops substantially on both backbones.
(3) \textit{Trajectory filtering effectively prevents biased and low-quality patterns from being distilled during fine-tuning.} Removing it drops \model from 4.19 to 4.03.

\section{Analysis}
We analyze the effectiveness of the extrinsic evaluation, examine failure modes in trajectory synthesis, and present efficiency comparisons alongside qualitative case studies.

\begin{figure}[t!]
  \centering
  \includegraphics[width=\linewidth]{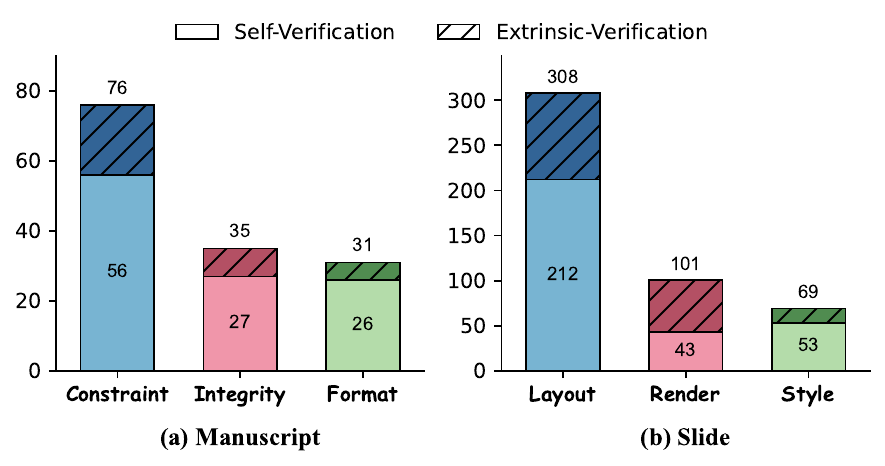}
  \caption{Distribution of defects identified by self-verification and extrinsic verification for manuscripts (left) and slides (right), respectively.}
  \label{fig:oversight}
\end{figure}

\subsection{Effect of Extrinsic Verification}

\paragraph{Extrinsic verification improves trajectory synthesis by mitigating self-verification bias.}
To quantify its impact, we train two variants on 300 trajectories sampled from the same set of tasks, with and without extrinsic verification during trajectory synthesis.
As shown in Table~\ref{tab:process_oversight}, adding extrinsic verification yields a 67\% larger gain in Avg. (0.20 vs. 0.12) than fine-tuning alone.
This indicates that, even with environment-grounded observations, revision signals produced solely within the agent's own trajectory state can be biased, leading to suboptimal refinements being distilled during learning.

\paragraph{Extrinsic verification mitigates self-verification bias by strengthening defect-triggered revision signals.}
We categorize reflection-triggered defects into three manuscript types: \textit{integrity} (e.g., missing asset references), \textit{constraint} (e.g., mismatched slide count), and \textit{format} (e.g., invalid markup); and three slide types: \textit{layout} (e.g., overlap), \textit{render} (e.g., blank slides), and \textit{style} (e.g., low contrast).
Figure~\ref{fig:oversight} compares defects identified on the same 300 trajectories under self-verification versus extrinsic verification.
Extrinsic verification consistently yields more defect detections across categories, with the largest gaps on slides (e.g., 308 vs.\ 212 for \textit{layout} and 101 vs.\ 43 for \textit{render}). 
This pattern indicates a systematic failure in self-verification: when verification is conducted within the generating trajectory state, the agent tends to rationalize defects, producing biased judgment \citep{jiang2025self, stechly2024self}. By decoupling verification from the agent's own trajectory state, extrinsic verification mitigates this bias and provides stronger signals to trigger corrective revisions during synthesis.

\begin{figure}[t!]
  \centering
  \includegraphics[width=\linewidth]{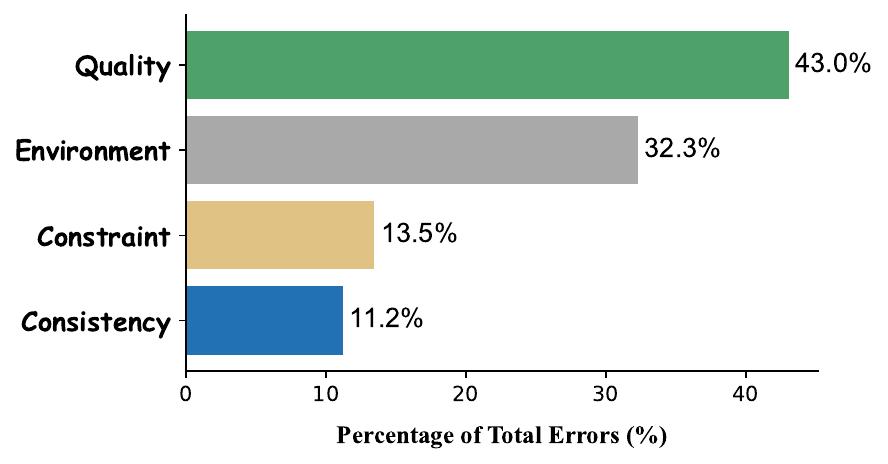}
  \caption{Failure distribution in synthesized trajectories before filtering}
  \label{fig:error_analysis}
\end{figure}

\begin{figure}[t!]
  \centering
  \includegraphics[width=\linewidth]{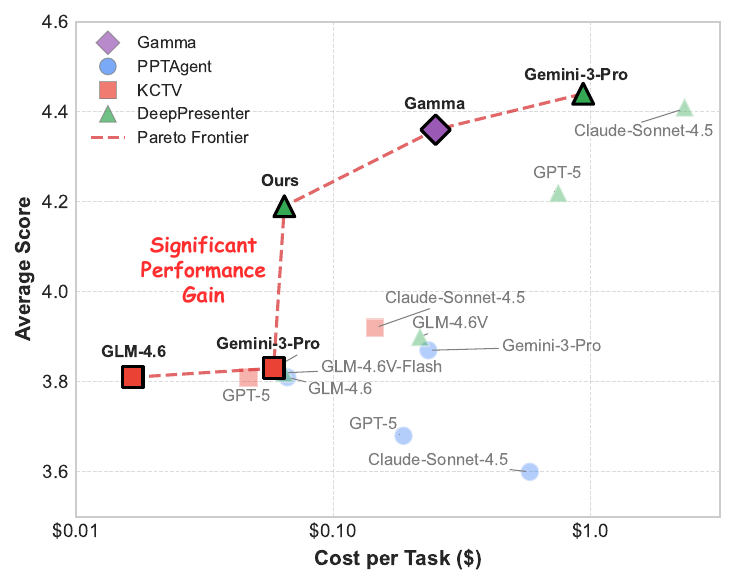}
  \caption{Performance vs. Price scatter plot with Pareto frontier representation. Different colors represent different frameworks}
  \label{fig:efficiency}
\end{figure}

\subsection{Trajectory Failure Analysis}

Following the categories in Section~\ref{sec:filter}, we analyze failures in synthesized trajectories before filtering (Figure~\ref{fig:error_analysis}). \textit{Quality} errors are most prevalent (43.0\%), underscoring the difficulty of sustaining high standards under free-form generation. \textit{Environment} failures are also common (32.3\%), reflecting long-horizon fragility from context overflow and infrastructure disruptions. The remaining cases include \textit{Constraint} violations (13.5\%) and \textit{Consistency} errors (11.2\%), which are less frequent but still non-negligible.

\begin{figure*}[ht!]
  \centering
  \includegraphics[width=\linewidth]{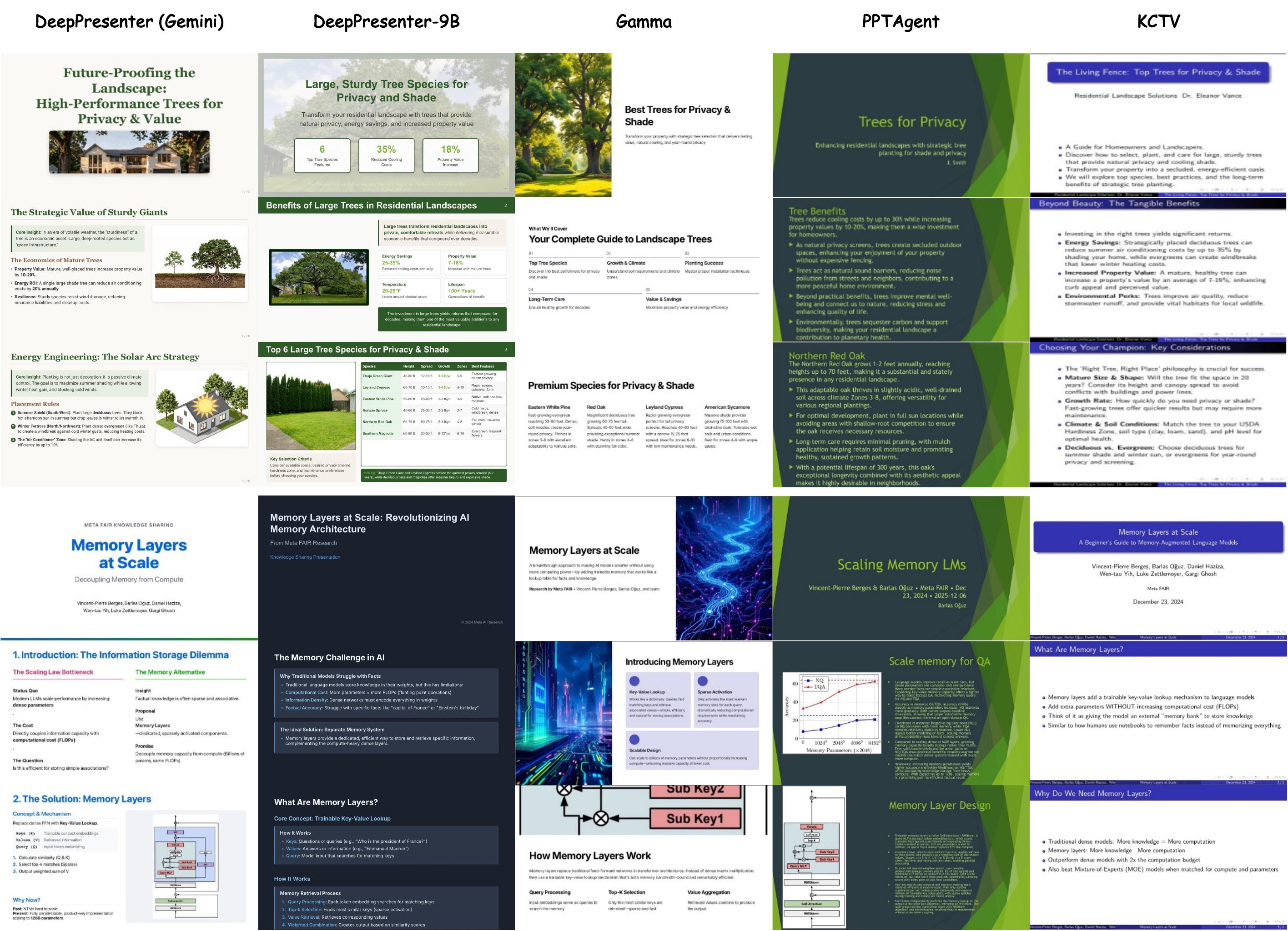}
  \caption{Qualitative comparison of presentations generated by different methods. \method under Gemini-3-Pro and \model produce high-quality slides with styles that resonate with the topic. Baselines rely on document-embedded or AI-generated images with template-based generation, producing text-heavy outputs and misaligned visual themes.}
  \label{fig:casestudy}
\end{figure*}

\subsection{Efficiency Analysis}
Figure~\ref{fig:efficiency} presents the cost-performance\footnote{The cost of fine-tuned models is estimated using the per-token pricing of their base models on OpenRouter.} trade-off across frameworks and models.
(1) \textit{\model advances the Pareto frontier, significantly outperforming the prior frontier point at comparable cost.}
Compared to KCTV + Gemini-3-Pro (3.83), \model achieves 4.19 at a similar price, a significant improvement in cost-efficiency.
(2) \textit{\method establishes a new upper bound for presentation generation, surpassing the previous best system Gamma.}
With an average score of 4.44 versus Gamma's 4.36, \method delivers the strongest result in our evaluation.

Notably, baseline frameworks exhibit flat performance across backbone models, whereas \method demonstrates substantial variation (3.82 to 4.44). This pattern is consistent with baselines being limited by their fixed pipelines, while \method can better leverage stronger model capacity.

\subsection{Case Study}
We present qualitative examples in Figure~\ref{fig:casestudy}. (1) \textit{\method produces visually rich slides through diverse asset sources, while baselines tend to yield text-heavy outputs.} Gamma includes more imagery than academic baselines. However, it relies heavily on AI-generated images and often mishandles figures embedded in source documents (e.g., inappropriate scaling of architectural diagrams). Open-source baselines rarely retrieve or create supporting visuals, resulting in predominantly textual content. (2) \textit{\method generates visual themes that resonate with content, whereas baselines rely on fixed templates.} For example, \method employs green tones for environmental topics and minimalist layouts for academic presentations, while baseline methods exhibit limited topical alignment due to template-driven generation.

\section{Related Work}

Presentation generation has attracted increasing attention due to its practical value for information delivery.
Before the emergence of large language models, presentation generation was primarily formulated as a document summarization task.
These approaches employed extractive summarization to select salient sentences using neural networks~\citep{hu2014ppsgen, Fu_Wang_McDuff_Song_2022, sun2021d2s} or phrase-based methods~\citep{wang2017phrase}.
However, the limited reasoning capabilities of pre-LLM models constrained their ability to handle diverse user intents and produce visually engaging outputs.

The emergence of LLMs has shifted the paradigm toward agent-based approaches that leverage stronger reasoning and generalization capabilities.
Recent work explores multi-agent collaboration for content extraction and layout planning~\citep{liang2025slidegen,yang2025auto,xu2025pregenie,ge2025autopresent,cachola-etal-2024-knowledge}, aesthetic-aware generation~\citep{liu2025presenting}, as well as slide understanding and editing~\citep{jung2025talk,huang2025pptbench,zheng2025pptagent,zeng2025slidetailor}.
However, these approaches often focus on predefined workflows and fixed templates, limiting adaptation to user intent and iterative refinement with environmental feedback.

Compared with previous methods, \method formulates presentation generation as an autonomous exploration and collaboration process between two specialized agents. The Researcher-Presenter decomposition enables adaptive planning based on task complexity, while environment-grounded reflection allows agents to verify and refine artifacts through rendered slides and file system states \citep{tang2025realcritic,jiang2025self}.

\section{Conclusion}

In this work, we propose \method, an agentic framework for presentation generation in which agents plan autonomously and adapt to diverse user intents. Our framework grounds self-reflection in perceptual artifact states from environmental observations, enabling agents to iteratively identify and fix post-render defects. We further train \model on trajectories synthesized with extrinsic verification, which mitigates self-verification bias and strengthens reflective behaviors. Results show that \method achieves state-of-the-art performance, while \model remains competitive at substantially lower cost.

\section*{Acknowledgments} We sincerely thank the reviewers for their insightful comments and valuable suggestions. This work was supported by Beijing Natural Science Foundation (L243006), the Natural Science Foundation of China (No. 62476265, 62306303, 62506354).

\section*{Limitations}

While \method demonstrates strong performance, several limitations remain.
First, \method relies on multi-step, tool-using rollouts, which increase inference cost and are sensitive to environment instability (e.g., context overflow and infrastructure failures) observed in our trajectory analysis.
Second, extrinsic verification is only used during trajectory synthesis. We do not employ an external critic at inference time, as critic-provided reflection signals can introduce reflection--action inconsistency and additional overhead. Future work can explore mitigating self-verification bias at inference time.

\bibliography{custom}

\clearpage

\appendix

\section{Detailed Analysis}

\subsection{Human Evaluation}
To address concerns about potential circularity introduced by LLM-as-judge evaluation, we conduct a small-scale human study to corroborate the automatic assessments. We recruited two graduate students majoring in computer science to evaluate 32 randomly sampled presentations from the test set. 
Following the evaluation dimensions in Section~\ref{sec:exp}, annotators rate Content and Style on a 1–5 Likert scale using the scoring criteria of \citet{zheng2025pptagent}, while Constraint satisfaction is verified via rule-based checks consistent with our evaluation protocol.
Evaluators were provided with rendered slide images and scored them independently. Table~\ref{tab:human_eval} reports the resulting ratings. Importantly, the relative ranking and overall trends under human judgment align with our automatic evaluation, suggesting that the observed improvements are not an artifact of relying solely on GPT-5 as the judge.

\subsection{Performance by Domain}
We analyze \method with Gemini-3-Pro across domains.
PersonaHub shows the strongest content (4.49) and style (4.49) scores, but relatively lower constraint satisfaction (4.38). This is likely because PersonaHub queries are synthesized by an LLM based on persona descriptions, resulting in more diverse and complex constraint specifications that are harder to follow.
arXiv achieves near-perfect constraint satisfaction (4.91) but the lowest content (3.84) and style (4.13) scores. The formal nature of academic presentations restricts visual diversity, and accurately conveying technical content requires deeper domain understanding.

\subsection{Tool Usage Analysis}

\begin{figure*}[t!]
  \centering
  \includegraphics[width=\linewidth]{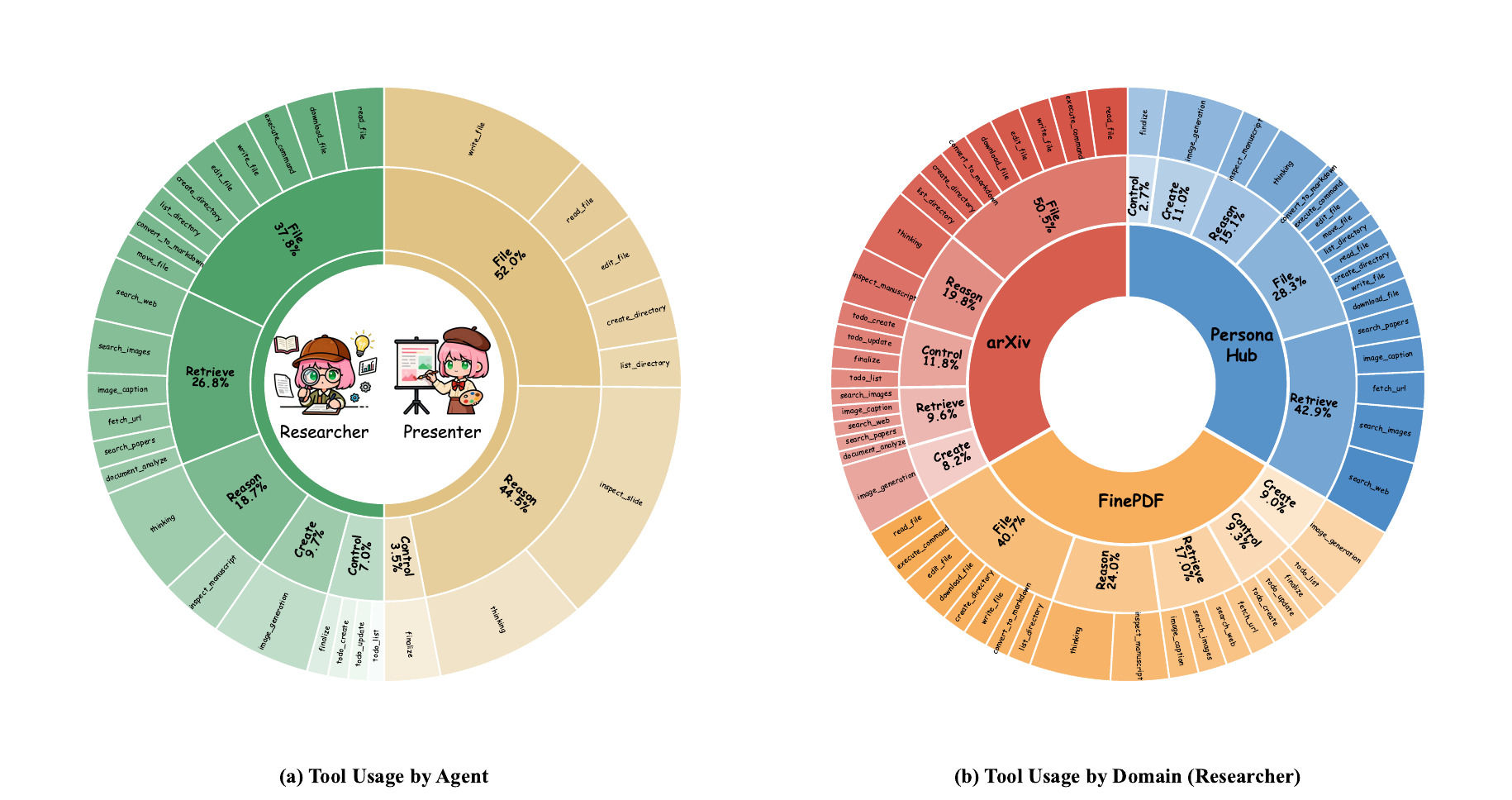}
  \caption{Tool usage analysis. (a) Distribution of tool invocations by agent role. (b) Tool usage patterns of \researcher across different domains.}
  \label{fig:tool_usage}
\end{figure*}

We analyze tool invocation patterns across agents and domains, as shown in Figure~\ref{fig:tool_usage}.
For agent roles (Figure~\ref{fig:tool_usage}a), \researcher and \presenter exhibit distinct tool preferences aligned with their responsibilities.
\researcher relies heavily on Retrieve tools for information gathering, while \presenter focuses on File operations and Reason tools for iterative slide editing and reflection.
This specialization validates our dual-agent design, where each agent develops tool usage patterns tailored to its role.

Across domains (Figure~\ref{fig:tool_usage}b), \researcher shows adaptable usage patterns reflecting task characteristics.
PersonaHub tasks exhibit significantly higher Retrieve usage, as persona-driven queries do not provide reference documents, requiring agents to search for relevant materials~\cite{yuan2025memsearcher}.
In contrast, arXiv and FinePDF tasks involve provided source documents, leading to higher File usage for document processing and lower reliance on retrieval.
Tool categories are detailed in Table~\ref{table:action_categories}.

\section{Dataset}

\subsection{Data Sources}
We collect presentation tasks from three sources to ensure diverse scenario coverage.
For academic presentations, we pair arXiv papers with requests that specify target audiences (beginners, intermediate learners, domain experts, or peer researchers) and corresponding scenarios (lectures, seminars, defenses, or conference talks).
For general educational topics, we sample English and Chinese PDF documents from FinePDFs-Edu~\citep{kydlicek2025finepdfs}, each accompanied by instructions to create a presentation based on the attachment.

\begin{table}[t!]
  \centering
  \resizebox{\columnwidth}{!}{
    \begin{tabular}{lcccc}
      \toprule
      \textbf{Method} & \textbf{Cons.} & \textbf{Content} & \textbf{Style} & \textbf{Avg.} \\
      \midrule
      Gamma & 4.84 & 3.52 & 3.90 & 4.09 \\
      PPTAgent & 3.72 & 3.07 & 3.60 & 3.46 \\
      KCTV  & 4.41 & 2.84 & 3.19 & 3.48 \\
      \midrule
      DeepPresenter & 4.56 & 3.86 & 4.25 & 4.22 \\
      \bottomrule
    \end{tabular}
  }
  \caption{Human evaluation results on 32 randomly sampled presentations.}
  \label{tab:human_eval}
\end{table}

\begin{table}[t!]
    \centering
    \begin{tabular}{lcccc}
        \toprule
        \textbf{Domain} & \textbf{Cons.} & \textbf{Content} & \textbf{Style} & \textbf{Avg.} \\
        \midrule
        PersonaHub  & 4.38 & 4.49 & 4.49 & 4.45 \\
        arXiv       & 4.91 & 3.84 & 4.13 & 4.29 \\
        FinePDF & 4.94 & 4.21 & 4.38 & 4.51 \\
        \bottomrule
    \end{tabular}
    \caption{Domain performance breakdown. Cons. denotes constraint satisfaction.}
    \label{tab:domain}
\end{table}

For personalized scenarios, we leverage PersonaHub~\citep{ge2024personahub} and prompt Qwen3-235B-A22B~\citep{yang2025qwen3} to generate realistic presentation requests grounded in user personas.
We adopt two generation strategies: knowledge-grounded generation, which incorporates both persona descriptions and synthesized domain knowledge, and open-ended generation, which relies solely on persona characteristics.
The model is instructed to adopt the persona's perspective and select the appropriate language based on cultural background.
Generated queries undergo language filtering, semantic deduplication, and LLM-based quality control to remove low-quality or inappropriate samples.

\subsection{Constraint Augmentation}
To assess instruction-following capabilities, each task is augmented with verifiable constraints, including slide count, aspect ratio (widescreen 16:9, standard 4:3, or poster), and language.
These constraints are randomly assigned per task.
For automated verification, we parse generated PDFs and validate them against specified constraints using a rule-based system.
The constraint satisfaction score is computed as the proportion of constraints successfully met.

\subsection{Evaluation Set}
To facilitate replication, we disclose the composition of our 128-task evaluation split and statistics in Table~\ref{table:evalset}.

\begin{table}[t]
  \centering
  \resizebox{\linewidth}{!}{
  \begin{tabular}{lccc}
    \toprule
    \textbf{Dimension} & \textbf{Category} & \textbf{Count} & \textbf{Ratio (\%)} \\
    \midrule
    \multirow{2}{*}{Language} & English & 74 & 57.81 \\
     & Chinese & 54 & 42.19 \\
    \midrule
    \multirow{3}{*}{Source} & PersonaHub & 57 & 44.53 \\
     & FinePDFs & 38 & 29.69 \\
     & arXiv & 33 & 25.78 \\
    \midrule
    \multirow{4}{*}{Aspect Ratio}
     & 16:9 Widescreen & 42 & 32.81 \\
      & 4:3 Standard & 34 & 26.56 \\
     & A1 Poster & 4 & 3.12 \\
     & Free & 48 & 37.50 \\
    \midrule
    \multirow{3}{*}{Slide Count}
     & 11-20 & 26 & 20.31 \\
     & 1-10 & 36 & 28.12 \\
      & Free & 66 & 51.56 \\
    \midrule
    \multicolumn{2}{l}{\textbf{Total}} & \textbf{128} & \textbf{100.00} \\
    \bottomrule
  \end{tabular}}
  \caption{Evaluation set statistics across language, source, aspect ratio, and slide-count constraints. ``Free''
indicates no constraint is specified.}
  \label{table:evalset}
\end{table}

\begin{table}[t!]
\centering
\resizebox{\columnwidth}{!}{
\begin{tabular}{>{\centering\arraybackslash}m{2cm}>{\raggedright\arraybackslash}m{5cm}}
\toprule
\textbf{Category} & \textbf{Action} \\
\midrule
Retrieve & search\_web, search\_images, search\_papers, fetch\_url, get\_paper\_authors, get\_scholar\_details , document\_analyze, image\_caption \\
\midrule
File & convert\_to\_markdown, read\_file, write\_file, move\_file, edit\_file, download\_file, execute\_command,  create\_directory, list\_directory\\
\midrule
Reason & thinking, inspect\_slide, inspect\_manuscript \\
\midrule
Control & todo\_create, todo\_update, todo\_list, finalize \\
\midrule
Create & image\_generation \\
\bottomrule
\end{tabular}
}
\caption{Action Categories}
\label{table:action_categories}
\end{table}

\section{Agent Framework}
\label{appendix:agent}

Presentation creation requires interacting with heterogeneous resources beyond static web text, including search results, images, papers, and local files, as well as inspecting intermediate artifacts such as manuscripts and rendered slides. To support this, we organize our toolset into five categories (Table~\ref{table:action_categories}): \textit{Retrieve} for information gathering, \textit{File} for document manipulation, \textit{Reason} for inspection and reflection, \textit{Control} for task management, and \textit{Synthesis} for code execution and asset generation.

\paragraph{Inspection Tools.}
The \textit{Reason} category includes two inspection tools that enable environment-grounded reflection:
\begin{itemize}
  \item \texttt{inspect\_manuscript}: Parses the markdown manuscript and returns structured diagnostics, including the total slide count, detected content language, and validation results for referenced image assets. The tool checks whether each image path exists, flags external URLs that should be downloaded locally, identifies missing alt text, and warns about duplicate image usage.
  \item \texttt{inspect\_slide}: Renders an HTML slide into a pixel image using a headless browser and returns the image to the agent's visual context. The tool supports multiple aspect ratios (16:9 widescreen, 4:3 standard, A1 poster) and enables agents to perceive visual defects such as contrast issues and element overflow that are invisible at the code level.
\end{itemize}

Each task is executed as a sequence of reasoning-action-observation steps within a maximum context window of 50K tokens. To prevent context overflow, our system sends warning messages when the accumulated window length reaches 50\% and 80\% of the maximum capacity, allowing the agent to adjust its strategy accordingly.

\section{Prompts}

\subsection{Data Synthesis Prompts}

\begin{tcolorbox}[colback=blue!5!white, colframe=blue!75!black, breakable, title = {\textsc{Query Synthesis (PersonaHub)}}]

\ttfamily
\footnotesize

Generate a slide creation request based on the following information:\\

\{hint\}\\

User persona:\\
\{persona\}\\

\end{tcolorbox}

\begin{tcolorbox}[colback=blue!5!white, colframe=blue!75!black, breakable, title = {\textsc{Query Synthesis (PersonaHub-Detail)}}]

  \ttfamily
  \footnotesize

  Assume you are a user with the following characteristics:\\

  <persona>\\
  \{persona\}\\
  </persona>\\

  You want to create a slide presentation based on the following topic:\\

  <presentation\_topic>\\
  \{synthesized\_text\}\\
  </presentation\_topic>\\

  \{hint\}\\

  Please generate a slide creation request based on the above persona and topic.\\

\end{tcolorbox}

\subsection{Extrinsic Verification Prompts}

\begin{tcolorbox}[colback=blue!5!white, colframe=blue!75!black, breakable, title = {\textsc{Extrinsic Verification for Researcher Agent}}]

  \ttfamily
  \footnotesize

  You are a professional slide content reviewer responsible for checking slide content for issues based on specified dimensions.\\

  \textbf{Review Dimensions}\\
  Your review authority is strictly limited to the following dimensions:\\
  $\bullet$ Image path compliance: Check whether local paths are used and whether captions are included\\
  $\bullet$ Language selection compliance: Check whether the document is written in the correct language\\
  $\bullet$ Language consistency: Check for unnecessary mixing of Chinese and English, or inconsistent style\\
  $\bullet$ Language correctness: Check for grammatical errors, spelling mistakes/typos\\
  $\bullet$ Tool-returned warnings: Check warning messages and evaluate their impact on user understanding\\

  \textbf{Problem Description Standards}\\
  When finding issues, use first person:\\
  $\bullet$ Problem Location: ``I noticed on this page...'' / ``The tool detected...''\\
  $\bullet$ Improvement Plan: ``I will...''\\

  \textbf{Return Format (strict JSON)}\\
  \{``severity'': <0-3 integer>, ``thought'': ``<analysis, less than 30 words>''\}\\

\end{tcolorbox}

\begin{tcolorbox}[colback=blue!5!white, colframe=blue!75!black, breakable, title = {\textsc{Extrinsic Verification for Presenter Agent}}]

  \ttfamily
  \footnotesize

  You are a professional slide design reviewer, responsible for analyzing the visual design and readability of HTML slides generated by another Design Agent and providing specific improvement guidelines.\\

  \textbf{Review Dimensions}\\
  \textit{1. Readability}\\
  $\bullet$ Whether the contrast between text and background is too low, causing reading difficulties\\
  $\bullet$ Whether fonts and images render properly\\
  $\bullet$ Whether text elements are obscured or overflow\\

  \textit{2. Aesthetics}\\
  $\bullet$ Whether similar elements maintain consistent alignment\\
  $\bullet$ Whether color schemes, visual hierarchy, and layout cause visual confusion\\
  $\bullet$ You should only check if images display correctly, not their aesthetics or watermarks\\

  \textbf{Problem Description Standards}\\
  When finding issues, use first person: ``I noticed on this slide...'' $\rightarrow$ ``This will cause...'' $\rightarrow$ ``I will...''\\

  \textbf{Return Format (strict JSON)}\\
  \{``severity'': <0-3 integer>, ``thought'': ``<analysis and improvement actions>''\}\\

\end{tcolorbox}

\subsection{Agent System Prompts}

\begin{tcolorbox}[colback=blue!5!white, colframe=blue!75!black, breakable, title = {\textsc{Researcher Agent System Prompt}}]

\ttfamily
\footnotesize

You are a professional presentation content expert capable of leveraging various tools for deep and comprehensive information retrieval and collection based on user requirements, then analyzing and highly distilling the information to create high-quality slide content that embodies ``Information Aesthetics''\\

\textbf{Task Instructions}\\
$\bullet$ Based on user requirements and their underlying logic, conduct systematic and comprehensive information research, and construct a slide framework with strong narrative tension\\
$\bullet$ After fully completing information collection and organization, organize visual materials guided by information value and content logic\\
$\bullet$ Write the manuscript in Markdown format: Use \texttt{---} for page separation; images must be downloaded locally and referenced via relative paths\\
$\bullet$ Upon completion, call \texttt{finalize} with the manuscript path as the parameter\\

\textbf{Important Notes}\\
$\bullet$ Use the same language as the user's instructions for manuscript generation\\
$\bullet$ Your task is limited to manuscript writing and material collection/creation; do not involve slide layout and design work\\
$\bullet$ Leverage `thinking` to reflect on the current state and next steps, and execute strictly\\
$\bullet$ You are not allowed to interact with the user; all information must be obtained through retrieval and tools\\

\end{tcolorbox}

\begin{tcolorbox}[colback=blue!5!white, colframe=blue!75!black, breakable, title = {\textsc{Presenter Agent System Prompt}}]

  \ttfamily
  \footnotesize

  You are a professional slide visual design expert, skilled in creating fixed-layout slide designs using HTML/CSS. Your core competency is faithfully transforming manuscripts into visually balanced, overlap-free, high-quality slides suitable for projection display, making full use of all available materials.\\

  \textbf{Task Description}\\
  $\bullet$ Deeply analyze the manuscript and develop a ``slide master'' style design plan (including color scheme, fonts, grid system, font size specifications)\\
  $\bullet$ Based on the design plan and manuscript content, generate high-quality HTML files page by page\\
  $\bullet$ After generating all slides, call the \texttt{finalize} tool to return the slides folder and end the workflow\\

  \textbf{Important Notes}\\
  $\bullet$ Mandatory Fixed Dimensions: Strictly lock body/html to specified dimensions (e.g., 16:9, 1280px x 720px), and set \texttt{overflow: hidden}\\
  $\bullet$ Use the same language as the user's instructions for thinking and working\\
  $\bullet$ Thinking: You can use `thinking` to reflect on the current state and plan next steps, then execute strictly\\

\end{tcolorbox}

\end{document}